\documentclass{article}
\usepackage{spconf,amsmath,graphicx}

\usepackage{color}

\title{Few-Shot Classification in Unseen Domains\\
by Episodic Meta-Learning Across Visual Domains
}
\name{Yuan-Chia Cheng$^{\star}$ \qquad Ci-Siang Lin$^{\star \dagger}$ \qquad Fu-En Yang$^{\star \dagger}$ \qquad Yu-Chiang Frank Wang$^{\star \dagger}$}
  
\address{$^{\star}$ Graduate Institute of Communication Engineering, National Taiwan University, Taiwan \\
  $^{\dagger}$ ASUS Intelligent Cloud Services, Taiwan}

\begin{document}
%
\maketitle
\begin{abstract}
  Few-shot classification aims to carry out classification given only few labeled examples for the categories of interest. Though several approaches have been proposed, most existing few-shot learning (FSL) models assume that base and novel classes are drawn from the same data domain. When it comes to recognizing novel-class data in an unseen domain, this becomes an even more challenging task of domain generalized few-shot classification. In this paper, we present a unique learning framework for domain-generalized few-shot classification, where base classes are from homogeneous multiple source domains, while novel classes to be recognized are from target domains which are not seen during training. By advancing meta-learning strategies, our learning framework exploits data across multiple source domains to capture domain-invariant features, with FSL ability introduced by metric-learning based mechanisms across support and query data. We conduct extensive experiments to verify the effectiveness of our proposed learning framework and show learning from small yet homogeneous source data is able to perform preferably against learning from large-scale one. Moreover, we provide insights into choices of backbone models for domain-generalized few-shot classification.
\end{abstract}
\section{Introduction}

\begin{figure*}[t]
    \centering
    \includegraphics[width=0.7\linewidth]{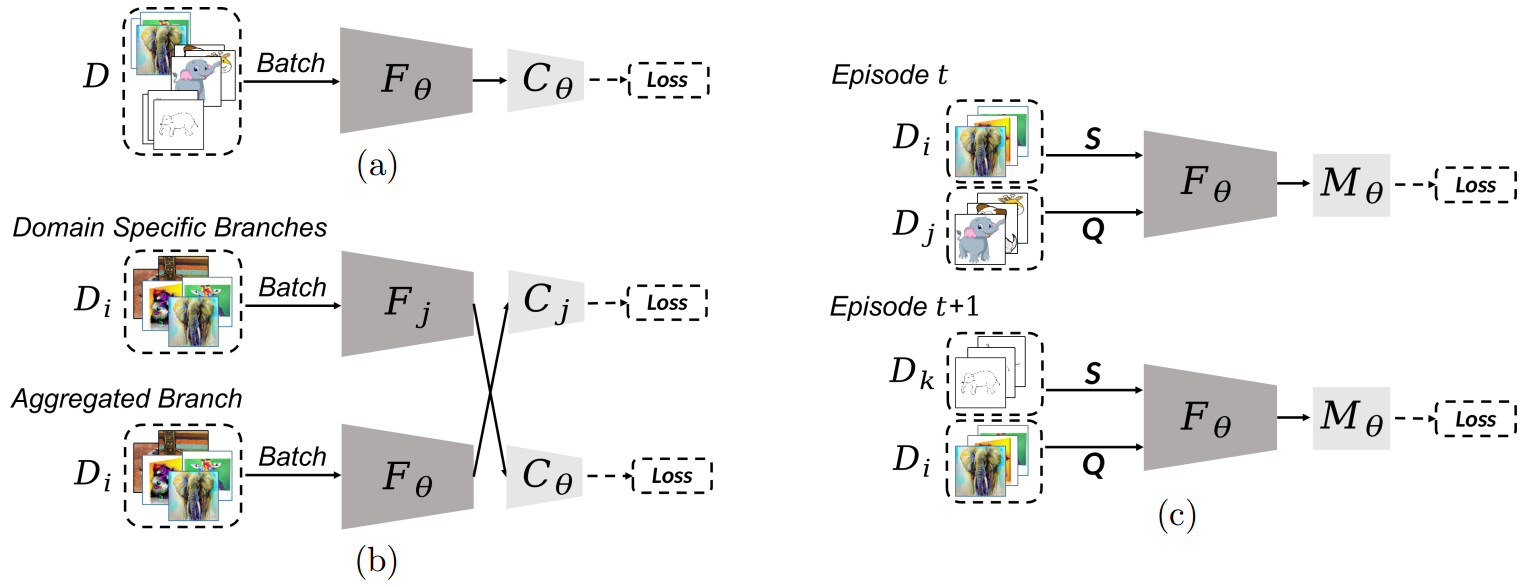}
    \caption{\textbf{Standard classification, domain-generalized classification, and domain-generalized few-shot classification.} 
    (a)~Naive baseline: do not distinguish between different source domains, and simply train feature encoder \(F_{\theta}\) and classifier \(C_{\theta}\) using data from source domains. (b) Episodic-learning for domain generalization~\cite{li2019episodic}: with feature encoder \(F_{j}\) and classifier \(C_{j}\) pre-trained on each source domain \(D_{j}\), train \(F_{\theta}\) and \(C_{\theta}\) via the associate paths to recognize test data in unseen domains (sharing the same label set). 
    (c) Cross-domain episodic meta-learning (x-EML): \(F_{\theta}\) and \(M_{\theta}\) (metric learner) are trained in $N$-way $K$-shot fashions, with support \(\mathbf{S}\) and query \(\mathbf{Q}\) sampled from distinct domains. \(F_{\theta}\) and \(M_{\theta}\) can thus be applied to recognize novel classes in unseen domains.}
    \label{fig:fig_1}
\end{figure*}

Recent development of deep learning technologies brings significant improvements on a variety of real-world applications. However, its high expenditure of collecting and labeling data generally limits its feasibility in real-world scenarios. To address this challenge, \emph{few-shot learning} (FSL) has been presented, and few-shot classification has been among active research topics in computer vision~\cite{finn2017model, finn2018probabilistic, nichol2018firstorder, hariharan2017low, antoniou2017data, wang2018low, vinyals2016matching, snell2017prototypical, sung2018learning}. Among FSL approaches, metric-based meta-learning models~\cite{vinyals2016matching, snell2017prototypical, sung2018learning} draw most attention by its simplicity and generalization ability to recognize novel classes.

Although few-shot classification has been extensively studied over the past few years, most existing methods assume that base and novel classes are both presented in the same data domain. Thus, significant performance drops would be expected under the domain generalization setting, i.e., novel classes are presented in an unseen domain, as pointed out in \cite{guo2019new, chen2019closer}. For the setting of \emph{domain generalization}, existing solutions and their goals are to utilize training data observed from multiple source domains, aiming at learning the model which can be generalized to unseen target domains for recognition tasks. To tackle this unseen domain shift issue, a plethora of approaches \cite{blanchard2011generalizing, muandet2013domain, li2017deeper, li2018domain, balaji2018metareg, li2018learning, li2019episodic, li2019feature, song2019generalizable} are proposed, which target at various applications such as image classification and person re-identification. Nevertheless, few efforts have been put into the domain-generalized few-shot classification. As a consequence, how to model and handle the unknown \emph{domain shift} remains challenging if recognizing cross-domain novel-class data is desirable.

In this paper, we tackle the task of domain-generalized few-shot classification, which recognizes data of novel classes in unseen domains. The learning of our proposed framework requires training data from multiple homogeneous source domains. Note that \textit{homogeneous} source-domain data indicate training data of base classes sharing the \textit{same} label set. Based on metric-based meta-learning, the proposed learning framework aims to derive domain-invariant features by recognizing support and query data from distinct source domains, with such a learning mechanism which can be generalized to data in unseen domains.

The contributions of our work are highlighted below:
\begin{itemize}

\item We address the task of domain-generalized few-shot classification, which learns domain-invariant features with domain generalization abilities for recognizing image data of novel categories in unseen domains.

\item To learn image features with domain generalization and FSL guarantees, we present a meta-learning scheme which performs metric-learning across support and query data sampled from distinct data domains.

\item Extensive experiments are conducted to demonstrate that deeper backbones would not be desirable for recognizing data of unseen categories/domains. This observation is quite different from that of standard classification tasks under single domain settings.

\end{itemize}

\section{Proposed Method}

\subsection{Notations and problem formulation}
For the sake of completeness, we first define the notations used in this paper. To tackle domain-generalized few-shot classification, we follow \cite{zhao2019multi} and consider base classes from multiple homogeneous source domains \(D = \{D_{1}, D_{2}, ..., D_{N_S}\}\) ($N_S$ is the number of source domains observed), which share the same label space. The novel classes are in a target domain \(D^{\ast}\), which is not available during training. In our meta-learning scheme, \(D\) is utilized during \emph{meta-training}, and only the \emph{meta-testing} stage has access to \(D^{\ast}\). For each training episode in the meta-training stage, we consider a standard \(N\)-way \(K\)-shot classification by constructing a support set \(\mathbf{S} = \{(\mathbf{X_{s}}, \mathbf{Y_{s}})\}\) and a query set \(\mathbf{Q} = \{(\mathbf{X_{q}} \mathbf{Y_{q}})\}\), while $\mathbf{S}$ and $\mathbf{Q}$ are sampled from distinct domains of $D$. 
As for meta-testing, we only have the access to support and query set data from \(D^{\ast}\). That is, our model observes the support set \(\mathbf{S}^{\ast} = \{(\mathbf{X}^{\ast}_{s}, \mathbf{Y}^{\ast}_{s})\}\) (also in a $N$-way $K$-shot fashion), and it is required to predict the labels for the query set \(\mathbf{Q}^{\ast} = \{\mathbf{X}^{\ast}_{q}\}\).

\begin{table*}[t]
    \centering
    \caption[c]{Performance comparisons on Omniglot, Mini-ImageNet, and CUB using ProtoNet as the backbone models. Note that Single Domain-S indicates the training of ProtoNet with support and query sets always from the same domain S, and x-EML* denotes the training of our model using support and query sets sampled from the same domain in each episode. The averaged accuracy is reported along with the 95\% confidence interval. \label{table:1}}
    
    \resizebox{0.9\linewidth}{!}{
        \begin{tabular}{l|c|c|c|c|c|c}
        & \multicolumn{2}{c|}{Omniglot} & \multicolumn{2}{c|}{Mini-ImageNet} & \multicolumn{2}{c}{CUB} \\ \cline{2-7}
        \textbf{Method} & 1-shot & 5-shot & 1-shot & 5-shot & \multicolumn{1}{c|}{1-shot} & 5-shot \\ \cline{1-7}
        \textbf{Single Domain-S} & $66.25\pm0.88$ & $86.35\pm0.75$ & $29.55\pm0.57$ & $42.34\pm0.71$ & $27.52\pm0.54$ & $38.27\pm0.61$ \\
        \textbf{Single Domain-C} & $66.73\pm0.89$ & $86.67\pm0.75$ & $31.86\pm0.61$ & $45.99\pm0.74$ & $28.72\pm0.54$ & $41.61\pm0.64$ \\
        \textbf{Single Domain-A} & $67.71\pm0.75$ & $87.02\pm0.71$ & $\bf 33.86\pm0.64$ & $48.77\pm0.70$ & $30.01\pm0.59$ & $42.77\pm0.64$ \\ \hline
        \bf Epi-FCR \cite{li2019episodic} & $67.14\pm0.82$ & $85.52\pm0.64$ & $28.82\pm0.62$ & $41.92\pm0.65$ & $29.15\pm0.55$ & $41.24\pm0.62$ \\
        \bf LFT \cite{tseng2020cross} & $64.22\pm0.93$ & $85.43\pm0.73$ & $31.06\pm0.59$ & $43.04\pm0.66$ & $28.81\pm0.58$ & $38.65\pm0.60$ \\ \hline
        \textbf{x-EML*} & $68.10\pm0.82$ & $88.55\pm0.66$ & $32.63\pm0.60$ & $48.38\pm0.73$ & $29.12\pm0.56$ & $42.90\pm0.66$ \\ \hline
        \textbf{x-EML (Ours)} & $\bf 69.90\pm0.81$ & $\bf 89.62\pm0.64$ & $33.35\pm0.64$ & $\bf 49.00\pm0.72$ & $\bf 30.11\pm0.57$ & $\bf 44.26\pm0.67$ \\
        \end{tabular}
    }
\end{table*}

\subsection{Cross-domain episodic meta-learning}
\label{ssec:cdeml}
Metric-learning based meta-learning has been widely applied for solving few-shot classification tasks \cite{vinyals2016matching, snell2017prototypical, sung2018learning}. However, they generally assume that both base and novel classes are from the same domain, and thus lack the ability to generalize across data domains. While one can naively include the data of base categories from multiple source domains \(D\) for training, as shown in Figure~\ref{fig:fig_1}(a), there is no guarantee that the learned model would be able to handle unseen target domain data. As depicted in Figure~\ref{fig:fig_1}(b), an episodic learning scheme was recently proposed by \cite{li2019episodic}, which utilizes local feature extractors and classifiers pre-trained in each data domain, guiding the learned global ones to generalize to unseen data domains. However, they require the training and test data (from seen and unseen domains) share exactly the same label space. Therefore, the episodic learning scheme would still not be able to recognize novel class in unseen domains directly.

To address domain-generalized few-shot classification, we propose Cross-Domain Episodic Meta-Learning (x-EML) for deriving domain-invariant yet semantics-discriminative representations, which can be applied to tackle novel class data in unseen target domains. As depicted in Figure~\ref{fig:fig_1}(c), the meta-training stage of our proposed framework follows the \(N\)-way \(K\)-shot scenario. To construct \(\mathbf{S}\) and \(\mathbf{Q}\) for each episode, we first randomly select \(N\) classes from a source domain \(D_{i}\) from \(D\), with \(K\) examples sampled for each class. On the other hand, to form \(\mathbf{Q}\), we sample \(m\) examples from a different source domain \(D_{j}\) for each of the \(N\) classes. Thus, for each episode in meta-training, we have a total of \(N \times K\) examples in \(\mathbf{S}\) along with \(N \times m\) examples in \(\mathbf{Q}\). Such support and query set samples are fed into the encoder \(F_{\theta}\) for feature extraction, followed by a proper metric-learning module \(M_{\theta}\) to predict class labels for each query example. We note that, for the metric-learning module \(M_{\theta}\), one can apply any existing metric-based algorithms or networks, comparing the similarity between \(F_{\theta}(X_{s})\) and \(F_{\theta}(x_{q})\) and assign labels (out of \(N\)) to \(F_{\theta}(x_{q})\). Moreover, we calculate the loss for the episode by loss function \(L\) and update the parameters \(\theta\) via conventional backpropagation and gradient descent strategies. Overall, the objective function for training the x-EML framework in each episode can be formulated as follows:

\begin{equation}
\begin{aligned}
\mathop{\arg\min}_{\theta} \sum_{(x_{q}, y_{q}) \in \mathbf{Q}} &L(y_{q}, M_{\theta}(\mathbf{Y}_{s}, F_{\theta}(\mathbf{X}_{s}), F_{\theta}(x_{q}))), \\
&\mathbf{S} \in D_{i}, \mathbf{Q} \in D_{j}, i \neq j.
\end{aligned}
\end{equation}

The unique characteristic of our proposed x-EML framework lies in that \(\mathbf{S}\) and \(\mathbf{Q}\) are sampled from different source domains but share the same label space. With such a proposed meta-learning scheme, we would derive semantics-discriminative yet domain-generalized features which serve the purposes of few-shot learning in unseen target domains. The effectiveness of our proposed model would be verified in the following section.

\section{Experiments}

\begin{table*}[t]
    \centering
    \caption[c]{Comparisons of FSL models learning from homogeneous source domains and a single large-scale dataset (with ProtoNet as the backbone). Note that * indicates \emph{in-domain} FSL rather than domain generalized FSL. \label{table:3}}
    
    \resizebox{0.95\linewidth}{!}{
         \begin{tabular}{l|c|c|c|c|c|c|c|c|c}
           & Training Domain (Dataset Size) & Omniglot & CUB & Texture & Flower & Aircraft & QuickDraw & Traffic & MSCOCO \\ \cline{1-10}
        
        {\bf MetaDataset-I} & ILSVRC + 7 target domains (37M) & 79.56* & 67.01* & 65.18* & 86.85* & 71.14* & 64.88* & 46.48 & 39.87 \\
        {\bf MetaDataset-II} & ILSVRC (1M) & 59.98 & $\mathbf{68.79}$ & $66.56$ & 85.27 & 53.10 & 48.96 & 47.12 & 41.00 \\
        {\bf x-EML} & PACS (10k) & $\mathbf{61.50}$ & 66.63 & $\mathbf{73.64}$ & $\mathbf{86.92}$ & $\mathbf{56.08}$ & $\mathbf{55.12}$ & $\mathbf{50.59}$ & $\mathbf{49.13}$ \\
        
        \end{tabular}
    }
\end{table*}

\subsection{Datasets}

To tackle domain-generalized few-shot classification, we apply the benchmark PACS dataset \cite{li2017deeper} as the source domain training data, which consists of four homogeneous data domains of Photo, Art-Painting, Cartoon and Sketch, sharing identical 7 categories with 9991 images in total. In particular, Photo (P) domain is excluded from training our model in this work; this is to ensure the domain shift between PACS and target domains like Mini-ImageNet and CUB. As for the target domains, we consider Omniglot~\cite{lake2011one}, Mini-ImageNet~\cite{Ravi2017OptimizationAA} and CUB~\cite{WahCUB_200_2011}. Omniglot is a hand-written character dataset of 50 alphabets containing 1623 images for each alphabet. Mini-ImageNet consists of a subset of 100 classes and includes 600 images for each class. CUB, a bird species fine-grained dataset, holds 200 classes and 11,788 images in total. Since the target-domain data are to be recognized, we do not split them into separate subsets as~\cite{vinyals2016matching, Ravi2017OptimizationAA, hilliard2018few} did. Instead, we take each target-domain dataset for testing purposes.

We also adopt MetaDataset~\cite{Triantafillou2020Meta-Dataset:} as target domains of interest. MetaDataset is a benchmark for few-shot learning encompassing 10 existing image datasets. We follow the training, validation and test splits released by~\cite{Triantafillou2020Meta-Dataset:}.

\subsection{Implementation details}
In our work, we employ the most commonly used metric-based deep learning models---ProtoNet~\cite{snell2017prototypical}. To achieve fair comparisons, all feature encoders consist of 5 ConvBlocks, denoted as Conv-5, and the channel size of ConvBlock is fixed to 512. All images are resized to 64\(\times\)64 as inputs while no data augmentation is applied, and we perform training via the SGD optimizer with a learning rate of 0.0005.

As for LFT~\cite{tseng2020cross} and Epi-FCR~\cite{li2019episodic}, we re-train the models and report the results. In addition, since Epi-FCR is not originally designed for few-shot classification, we apply ProtoNet prediction strategy during evaluation.

All experiments in this paper, except for ones in Table~\ref{table:3}, are conducted in standard \(N\)-way \(K\)-shot classification. That is, following \cite{chen2019closer}, we randomly sample \(N\) classes from the target domain, with \(K\) and 16 images sampled for each class to form support and query sets, respectively. We present the quantitative results by averaging the accuracies over 1000 runs for Omniglot \cite{snell2017prototypical}, and 600 runs for Mini-ImageNet, CUB~\cite{chen2019closer} and MetaDataset.

\subsection{Quantitative evaluation}

We first present few-shot classification results on unseen target domains in Tables~\ref{table:1}. For comparison purposes, we first consider baseline methods of using training data of a single domain in PACS for training the few-shot learning models (i.e., Single Domain-S, C, A in the above tables). For state-of-the-art domain-generalized FSL methods, here we consider Epi-FCR~\cite{li2019episodic} and LFT~\cite{tseng2020cross} as competitors.

From the results shown in Tables~\ref{table:1}, we observe that our x-EML was able to produce promising recognition results on novel classes in unseen target domains. We note that, in order to verify the effectiveness of cross-domain meta-learning strategy presented in x-EML, we considered a controlled version (denoted as x-EML*) which collected support and query sets from the same domain (i.e., \(D_{i} = D_{j}\)) in each episode during training. Although satisfactory results were achieved by this controlled version, our full model x-EML achieved the best domain-generalized few-shot classification performances.

\subsection{Deeper backbones for better domain generalization?}

In~\cite{chen2019closer}, it is observed that a deeper model backbone is able to improve the performances in few-shot classification tasks under the \emph{single domain} setting. In this work, we are intrigued to observe if the same property holds under the domain-generalized few-shot classification tasks.

\begin{figure}[t]
    \centering
    \includegraphics[width=0.65\linewidth]{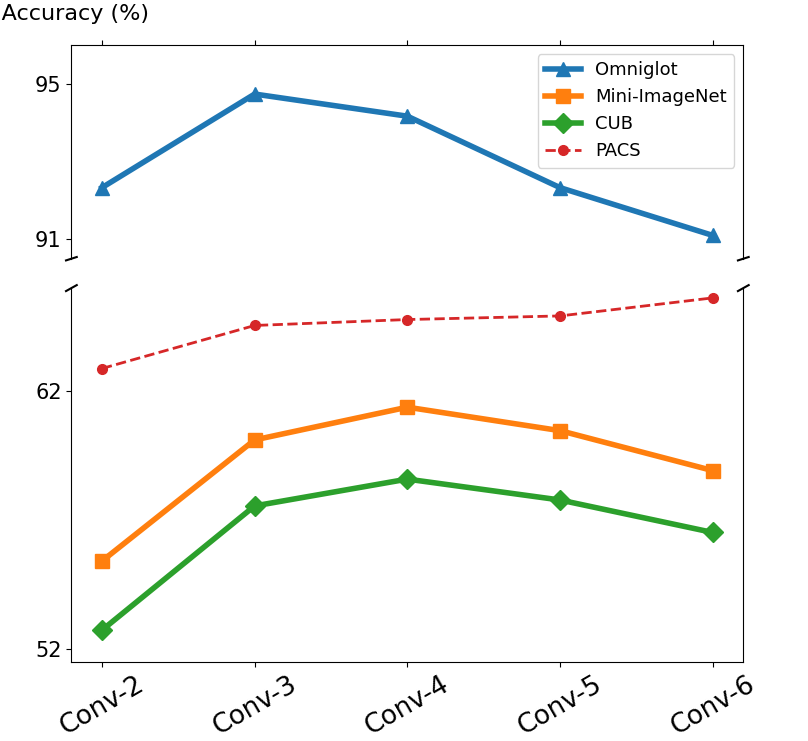}
    \caption[c]{Effects of backbones with varying depths on domain-generalized few-shot classification. We consider ProtoNet as the backbone of our x-EML framework, and Conv-\(n\) indicates \(n\) ConvBlocks. Note that x-EML is trained on PACS in 3-way 5-shot settings, with results presented on Omniglot, Mini-ImageNet, and CUB. It can be seen that, use of deeper backbones is not necessarily preferable for domain generalization due to possible overfitting of source domains.}
    \label{fig:fig_2}
\end{figure}

To assess this issue, we carry out extensive experiments by incrementally stacking ConvBlocks to build up feature encoder backbones. As the results shown in Figure~\ref{fig:fig_2}, we observe that deeper backbones did not necessarily ensure better generalization (e.g., Conv-3 and Conv-4 were preferable while the performances declined generally as the backbone grew deeper). This is very different from what was observed in the standard single-domain few-shot classification tasks (as concluded in~\cite{chen2019closer}). To be more precise, we see that features learned via deeper backbones tended to better model the training data and its domain (i.e., PACS), and hence limited the domain generalization capability. The above experiments verify and explain why in cross-domain meta-learning schemes, training models with deeper backbones would not necessarily increase the generalization ability.

\subsection{Learn from large-scale single-domain data or not?}

Despite promising results have been achieved by our proposed learning scheme, one might argue the learning from a single yet large-scale dataset would be sufficient. To clarify this issue, we adopt MetaDataset~\cite{Triantafillou2020Meta-Dataset:} and consider two FSL models for additional comparisons: MetaDataset-I and MetaDataset-II. MetaDataset-I trains FSL models using images from both ILSVRC (as the source domain) and 7 target domains; on the other hand, MetaDataset-II learns from only ILSVRC, and all target domain data are unseen during training. Note that we follow the experimental protocol in~\cite{Triantafillou2020Meta-Dataset:}.

As shown in Table~\ref{table:3}, we performed favorably against MetaDataset-I and II for domain generalized FSL. Although MetaDataset-I reported satisfactory results on first six target domains, those domains are observed during training and thus can not be viewed as domain generalized FSL. Overall, our proposed learning strategy would be preferable for FSL models to recognize novel classes in unseen target domains.

\section{Conclusions}


In this paper, we proposed a novel learning framework, Cross-Domain Episodic Meta-Learning (x-EML), to address domain-generalized few-shot classification tasks. To recognize novel image classes in unseen domains, our approach derives category-discriminative yet domain-invariant representations from multiple homogeneous source domains. Our experimental results quantitatively verify the effectiveness of our x-EML and its superiority over the baselines and the state-of-the-art methods. Moreover, we show that learning from multi-source domain data with a moderate size would be preferable, instead of learning from a large-scale one. In addition, we provide insights into the choice of backbone models, explaining why deeper backbones do not necessarily generalize across distinct data domains.

\section{Acknowledgement}

This work is supported in part by the Ministry of Science and Technology of Taiwan under grant MOST 110-2634-F-002-036. We also thank National Center for High-performance Computing (NCHC) for providing computational and storage resources.

\bibliographystyle{IEEEbib}
\bibliography{ref}

\begin{thebibliography}{10}

\bibitem{li2019episodic}
Da~Li, Jianshu Zhang, Yongxin Yang, Cong Liu, Yi-Zhe Song, and Timothy~M
  Hospedales,
\newblock ``Episodic training for domain generalization,''
\newblock in {\em ICCV}, 2019.

\bibitem{finn2017model}
Chelsea Finn, Pieter Abbeel, and Sergey Levine,
\newblock ``Model-agnostic meta-learning for fast adaptation of deep
  networks,''
\newblock in {\em ICML}, 2017.

\bibitem{finn2018probabilistic}
Chelsea Finn, Kelvin Xu, and Sergey Levine,
\newblock ``Probabilistic model-agnostic meta-learning,''
\newblock in {\em NeurIPS}, 2018.

\bibitem{nichol2018firstorder}
Alex Nichol, Joshua Achiam, and John Schulman,
\newblock ``On first-order meta-learning algorithms,''
\newblock {\em arXiv}, 2018.

\bibitem{hariharan2017low}
Bharath Hariharan and Ross Girshick,
\newblock ``Low-shot visual recognition by shrinking and hallucinating
  features,''
\newblock in {\em ICCV}, 2017.

\bibitem{antoniou2017data}
Antreas Antoniou, Amos Storkey, and Harrison Edwards,
\newblock ``Data augmentation generative adversarial networks,''
\newblock {\em arXiv}, 2017.

\bibitem{wang2018low}
Yu-Xiong Wang, Ross Girshick, Martial Hebert, and Bharath Hariharan,
\newblock ``Low-shot learning from imaginary data,''
\newblock in {\em CVPR}, 2018.

\bibitem{vinyals2016matching}
Oriol Vinyals, Charles Blundell, Timothy Lillicrap, Daan Wierstra, et~al.,
\newblock ``Matching networks for one shot learning,''
\newblock in {\em NeurIPS}, 2016.

\bibitem{snell2017prototypical}
Jake Snell, Kevin Swersky, and Richard Zemel,
\newblock ``Prototypical networks for few-shot learning,''
\newblock in {\em NeurIPS}, 2017.

\bibitem{sung2018learning}
Flood Sung, Yongxin Yang, Li~Zhang, Tao Xiang, Philip~HS Torr, and Timothy~M
  Hospedales,
\newblock ``Learning to compare: Relation network for few-shot learning,''
\newblock in {\em CVPR}, 2018.

\bibitem{guo2019new}
Yunhui Guo, Noel~CF Codella, Leonid Karlinsky, John~R Smith, Tajana Rosing, and
  Rogerio Feris,
\newblock ``A new benchmark for evaluation of cross-domain few-shot learning,''
\newblock {\em arXiv}, 2019.

\bibitem{chen2019closer}
Wei-Yu Chen, Yen-Cheng Liu, Zsolt Kira, Yu-Chiang~Frank Wang, and Jia-Bin
  Huang,
\newblock ``A closer look at few-shot classification,''
\newblock in {\em ICLR}, 2019.

\bibitem{blanchard2011generalizing}
Gilles Blanchard, Gyemin Lee, and Clayton Scott,
\newblock ``Generalizing from several related classification tasks to a new
  unlabeled sample,''
\newblock in {\em NeurIPS}, 2011.

\bibitem{muandet2013domain}
Krikamol Muandet, David Balduzzi, and Bernhard Sch{\"o}lkopf,
\newblock ``Domain generalization via invariant feature representation,''
\newblock in {\em ICML}, 2013.

\bibitem{li2017deeper}
Da~Li, Yongxin Yang, Yi-Zhe Song, and Timothy~M Hospedales,
\newblock ``Deeper, broader and artier domain generalization,''
\newblock in {\em ICCV}, 2017.

\bibitem{li2018domain}
Haoliang Li, Sinno Jialin~Pan, Shiqi Wang, and Alex~C Kot,
\newblock ``Domain generalization with adversarial feature learning,''
\newblock in {\em CVPR}, 2018.

\bibitem{balaji2018metareg}
Yogesh Balaji, Swami Sankaranarayanan, and Rama Chellappa,
\newblock ``Metareg: Towards domain generalization using meta-regularization,''
\newblock in {\em NeurIPS}, 2018.

\bibitem{li2018learning}
Da~Li, Yongxin Yang, Yi-Zhe Song, and Timothy~M Hospedales,
\newblock ``Learning to generalize: Meta-learning for domain generalization,''
\newblock in {\em AAAI}, 2018.

\bibitem{li2019feature}
Yiying Li, Yongxin Yang, Wei Zhou, and Timothy~M Hospedales,
\newblock ``Feature-critic networks for heterogeneous domain generalization,''
\newblock in {\em ICLR}, 2019.

\bibitem{song2019generalizable}
Jifei Song, Yongxin Yang, Yi-Zhe Song, Tao Xiang, and Timothy~M Hospedales,
\newblock ``Generalizable person re-identification by domain-invariant mapping
  network,''
\newblock in {\em CVPR}, 2019.

\bibitem{zhao2019multi}
Sicheng Zhao, Bo~Li, Xiangyu Yue, Yang Gu, Pengfei Xu, Runbo Hu, Hua Chai, and
  Kurt Keutzer,
\newblock ``Multi-source domain adaptation for semantic segmentation,''
\newblock in {\em NeurIPS}, 2019.

\bibitem{tseng2020cross}
Hung-Yu Tseng, Hsin-Ying Lee, Jia-Bin Huang, and Ming-Hsuan Yang,
\newblock ``Cross-domain few-shot classification via learned feature-wise
  transformation,''
\newblock in {\em ICLR}, 2020.

\bibitem{lake2011one}
Brenden Lake, Ruslan Salakhutdinov, Jason Gross, and Joshua Tenenbaum,
\newblock ``One shot learning of simple visual concepts,''
\newblock in {\em CogSci}, 2011.

\bibitem{Ravi2017OptimizationAA}
Sachin Ravi and Hugo Larochelle,
\newblock ``Optimization as a model for few-shot learning,''
\newblock in {\em ICLR}, 2017.

\bibitem{WahCUB_200_2011}
Catherine Wah, Steve Branson, Peter Welinder, Pietro Perona, and Serge
  Belongie,
\newblock ``The caltech-ucsd birds-200-2011 dataset,''
\newblock Tech. {R}ep., California Institute of Technology, 2011.

\bibitem{hilliard2018few}
Nathan Hilliard, Lawrence Phillips, Scott Howland, Art{\"e}m Yankov, Courtney~D
  Corley, and Nathan~O Hodas,
\newblock ``Few-shot learning with metric-agnostic conditional embeddings,''
\newblock {\em arXiv}, 2018.

\bibitem{Triantafillou2020Meta-Dataset:}
Eleni Triantafillou, Tyler Zhu, Vincent Dumoulin, Pascal Lamblin, Utku Evci,
  Kelvin Xu, Ross Goroshin, Carles Gelada, Kevin Swersky, Pierre-Antoine
  Manzagol, and Hugo Larochelle,
\newblock ``Meta-dataset: a dataset of datasets for learning to learn from few
  examples,''
\newblock in {\em ICLR}, 2020.

\end{thebibliography}

\end{document}